\documentclass[sigconf, balance=false]{acmart}

\AtBeginDocument{%
  \providecommand\BibTeX{{%
    \normalfont B\kern-0.5em{\scshape i\kern-0.25em b}\kern-0.8em\TeX}}}

\copyrightyear{2023} 
\acmYear{2023} 
\setcopyright{acmlicensed}
\acmBooktitle{International Conference on Multimedia Retrieval (ICMR '23), June 12--15, 2023, Thessaloniki, Greece}

\acmConference[ICMR '23]{International Conference on Multimedia Retrieval}{June 12--15, 2023}{Thessaloniki, Greece}
\acmPrice{15.00}
\acmDOI{10.1145/3591106.3592267}
\acmISBN{979-8-4007-0178-8/23/06}

\acmSubmissionID{5906}

\usepackage{algorithm}
\usepackage{algorithmic}
\usepackage{multirow}
\usepackage{colortbl}

\newcommand{\etal}{\textit{et al}. }
\newcommand{\ie}{\textit{i}.\textit{e}., }
\newcommand{\eg}{\textit{e}.\textit{g}. }

\definecolor{mygray}{gray}{0.95}

\begin{document}

\title[Multi-Label Meta Weighting for Long-Tailed Dynamic Scene Graph Generation]{Multi-Label Meta Weighting\\for Long-Tailed Dynamic Scene Graph Generation}

\author{Shuo Chen, Yingjun Du, Pascal Mettes, and Cees G. M. Snoek}
\affiliation{
  \institution{University of Amsterdam}
  \country{}
}


\begin{abstract}
  This paper investigates the problem of scene graph generation in videos with the aim of capturing semantic relations between subjects and objects in the form of $\langle$subject, predicate, object$\rangle$ triplets. 
  Recognizing the predicate between subject and object pairs is imbalanced and multi-label in nature, ranging from ubiquitous interactions such as spatial relationships (\eg \emph{in front of}) to rare interactions such as \emph{twisting}. In widely-used benchmarks such as Action Genome and VidOR, the imbalance ratio between the most and least frequent predicates reaches 3,218 and 3,408, respectively, surpassing even benchmarks specifically designed for long-tailed recognition. Due to the long-tailed distributions and label
  co-occurrences, recent state-of-the-art methods predominantly focus on the most frequently occurring predicate classes, ignoring those in the long tail. 
  In this paper, we analyze the limitations of current approaches for scene graph generation in videos and identify a one-to-one correspondence between predicate frequency and recall performance. To make the step towards unbiased scene graph generation in videos, we introduce a multi-label meta-learning framework to deal with the biased predicate distribution. 
  Our meta-learning framework learns a meta-weight network for each training sample over all possible label losses. We evaluate our approach on the Action Genome and VidOR benchmarks by building upon two current state-of-the-art methods for each benchmark. The experiments demonstrate that the multi-label meta-weight network improves the performance for predicates in the long tail without compromising performance for head classes, resulting in better overall performance and favorable generalizability.
  Code: \url{https://github.com/shanshuo/ML-MWN}.
\end{abstract}





\maketitle

\section{Introduction}
Scene graph generation in videos focuses on detecting and recognizing relationships between pairs of subjects and objects. 
The resulting dynamic scene graph is a directed graph whose nodes are objects with their relationships as edges in a video. 
Extracting such graphs from videos constitutes a highly challenging research problem~\cite{johnson2015image}, with broad applicability in multimedia and computer vision. Effectively capturing such structural-semantic information boosts downstream tasks such as captioning~\cite{show}, video retrieval~\cite{snoek2009concept}, visual question answering~\cite{antol2015vqa}, and numerous other visual-language tasks. 

\begin{figure}
    \centering
    \includegraphics[width=\linewidth]{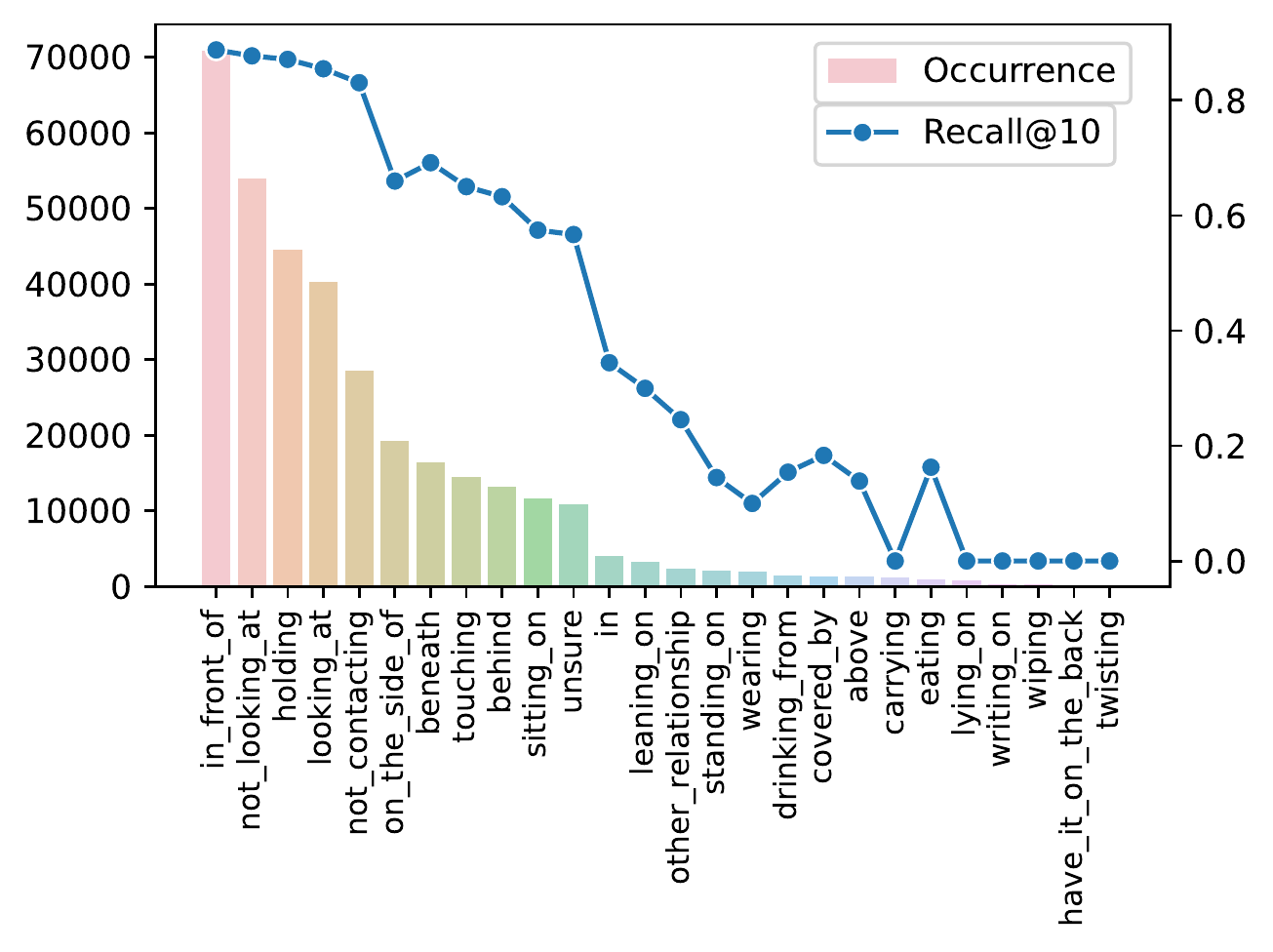}
    \caption{Long-tailed predicate occurrences vs. class-wise recall from STTran~\cite{cong2021spatial} on the Action Genome dataset~\cite{ji2020action}. The class-wise performance drops drastically, highlighting the importance of long-tailed dynamic scene graph generation.}
    \label{fig:ag}
\end{figure}

\begin{figure*}
    \centering
    \includegraphics[width=.9\linewidth]{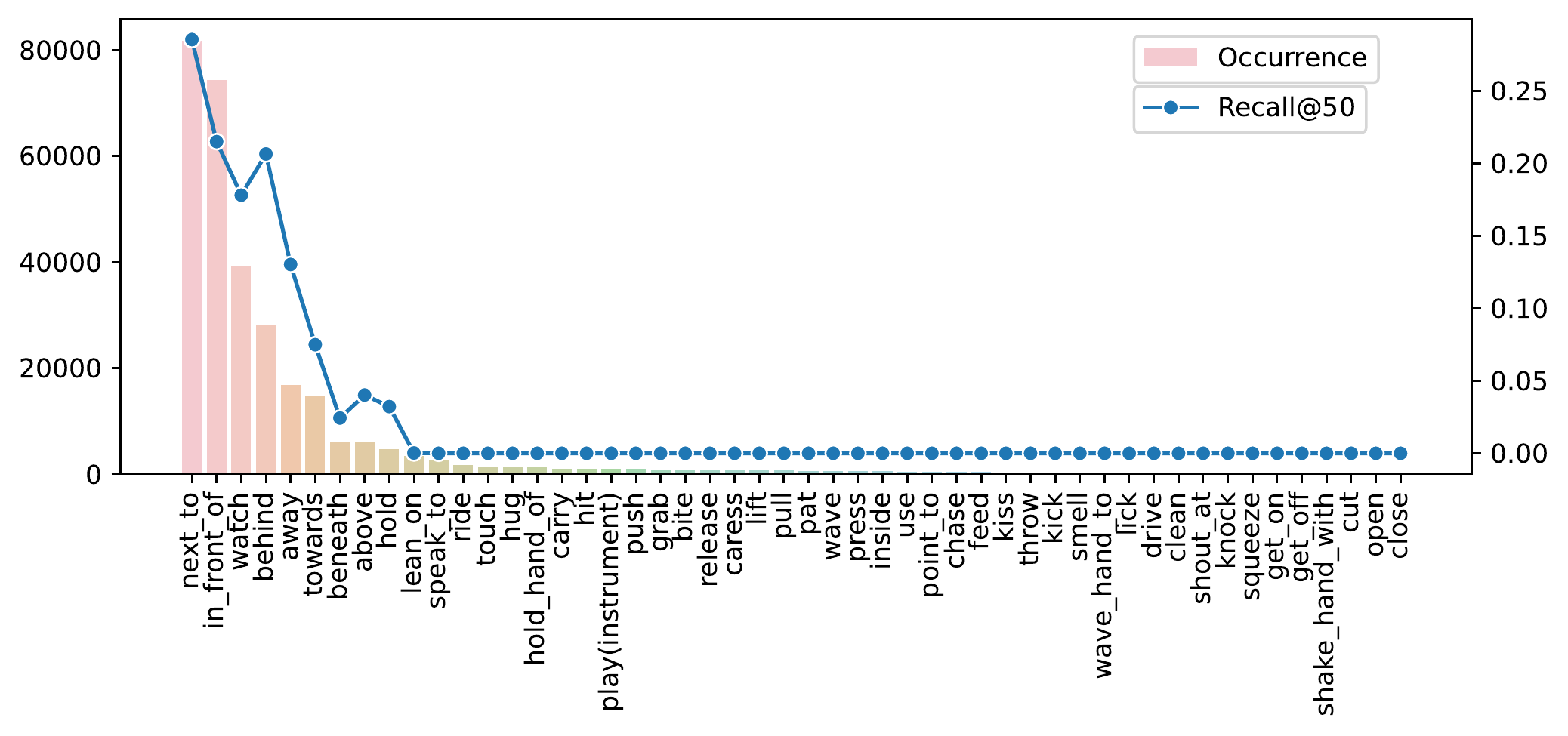}
    \caption{
    On the VidOR dataset~\cite{shang2019annotating}, the long-tailed distribution is even worse. We can observe that Social Fabric~\cite{chen2021social} ignores most predicates with limited samples.
    }
    \label{fig:vidor}
\end{figure*}

Current methods place a heavy emphasis on recognizing subject-to-object relationship categories. 
A leading approach to date involves extracting multi-modal features for relation instances, followed by either pooling the multi-modal features~\cite{qian2019video, su2020video, xie2020video} or learning a feature representation~\cite{chen2021social} to feed into the predicate classifier network.
Despite the strong focus on relation recognition, existing methods overlook the extremely long-tailed distribution of predicate classes. 
Figure~\ref{fig:ag} displays the recall per predicate class from STTran~\cite{cong2021spatial} and its corresponding occurrences on the Action Genome dataset.
This trend is even more pronounced on the VidOR dataset.
Figure~\ref{fig:vidor} illustrates the occurrence distribution vs. Recall@50 from Social Fabric~\cite{chen2021social} for the video relation detection task on the VidOR dataset, where a few head predicates dominate all other classes. This phenomenon has not been actively investigated, as the evaluation metrics do not penalize lower scores for predicates in the long tail. 
In light of these observations, this paper advocates for the development for scene graph generation methods in videos that effectively handle both common and rare predicates.

We introduce a meta-learning framework to address the long-tailed dynamic scene graph generation problem. 
Drawing inspiration from the concept of meta weighting~\cite{shu2019meta}, we propose a Multi-Label Meta Weight Network (ML-MWN) to learn meta weights across both examples and classes explicitly. 
These meta weights are, in turn, used to steer the downstream loss to optimize the parameters of the predicate classifier.
We adopt a meta-learning framework to optimize the ML-MWN parameters, where we compute each instance's per-class loss in a training batch and obtain a loss matrix. 
The loss matrix is fed into our ML-MWN, which outputs a weight matrix, with each row representing the weight vector for an instance's loss vector. 
We sample a meta-validation batch and use an unbiased meta-loss to guide the training of ML-MWN.
We adopt the inverse frequency binary cross-entropy loss as the meta-loss. 
Finally, we integrate our framework with existing methods to guide the predicate classification.

To evaluate our meta-learning framework, we employ two recent state-of-the-art methods~\cite{chen2021social,cong2021spatial}, one for the scene graph generation task on the Action Genome dataset and one for video relation detection on the VidOR dataset. 
We empirically demonstrate that our approach enhances predicate predictions for these recent methods across various evaluation metrics. 
Furthermore, we show that our framework improves the performance of long-tailed predicates without hampering the performance of more common classes. 
Our approach is generic and works on top of any scene graph generation method, ensuring broad applicability. 
We make the code available on \url{https://github.com/shanshuo/ML-MWN}.

In summary, our contributions are three-fold:

1. We investigate the long-tail issue in dynamic scene graph generation and analyze the limitations of existing methods.

2. We introduce a multi-label meta-learning framework to address the biased predicate class distribution.

3. We propose a Multi-Label Meta Weight Network (ML-MWN) to explicitly learn a weighting function, which demonstrates generalization ability performance on two benchmarks when plugged into two existing approaches,

\section{Related Works}
\label{sec:rel}

\paragraph{Dynamic scene graph generation.}
Scene graph generation was first pioneered in~\cite{johnson2015image} for image retrieval, and the task quickly gained further traction, as seen in \eg~\cite{xu2017scene, zellers2018neural, yang2018graph, liu2021fully, Tao2022Predicate}.
Recently, a number of papers have identified the long-tailed distribution in image scene graphs and focused on generating unbiased scene graphs~\cite{li2021bipartite, desai2021learning, yan2020pcpl, li2022ppdl, dong2022stacked, li2022dynamic}. We seek to bring the same problem to light in the video domain.
Ji~\etal~\cite{ji2020action} firstly extended scene graph generation to videos and introduced the Action Genome dataset.
A wide range of works have since proposed solutions to the problem~\cite{sun2019video,zheng2019relation,liu2020beyond, cao2021relation,xie2020video,su2020video,Kukleva_2020_CVPR,Sunkesula_mm_2020,chen2021social,gao2022classification}.
Recently, Li~\etal~\cite{li2022dynamic} proposed an anticipatory pre-training paradigm based on Transformer to model the temporal correlation of visual relationships.
Similarly, the VidOR dataset collected by Shang~\etal~\cite{shang2019annotating} is another popular benchmark.
Leading approaches generate proposals~\cite{chen2020interactivity} for individual objects on short video snippets, encode the proposals, predict a relation, and associate the relations over the entire video, \eg~\cite{qian2019video,su2020video,xie2020video}. Liu~\etal~\cite{liu2020beyond} generate the proposals using the sliding window way.
More recently, Gao~\etal~\cite{gao2022classification} proposed a classification-then-grounding framework, which can avoid the high influence of proposal quality on performance.
Chen~\etal~\cite{chen2021diagnosing} performed a series of analyses on video relation detection.
In this paper, we use STTran~\cite{cong2021spatial} and Social Fabric~\cite{chen2021social} to capture the relation feature and insert our multi-label meta-weight network on top.
Cong~\etal~\cite{cong2021spatial} proposed a spatial-temporal Transformer to capture the spatial context and temporal dependencies for a dynamic scene graph.
Moreover, Chen~\etal~\cite{chen2021social} proposed an encoding that represents a pair of object tubelets as a composition of interaction primitives. 
Both approaches provide competitive results and form a fruitful testbed for our meta-learning framework.

\paragraph{Multi-label long-tailed classification.}
Multi-label long-tailed recognition is a challenging problem that deals with sampling differences and biased label co-occurrences~\cite{zhang2021deep}.
A few works have studied this topic, with most solutions based on new loss formulations.
Specifically, Wu~\etal~\cite{wu2020distribution} proposed a distribution-based loss for multi-label long-tailed image recognition.
More recently, Tian~\etal~\cite{tian2022striking} proposed a hard-class mining loss for the semantic segmentation task by dynamically weighting the loss for each class based on instantaneous recall performance.
Inspired by these loss-based works, we utilize inverse frequency cross-entropy loss during our meta-learning process.

\paragraph{Meta learning for sample weighting.}
Ren~\etal~\cite{ren2018learning} pioneered the adoption of a meta learning framework to re-weight samples for imbalanced datasets.
Based on~\cite{ren2018learning}, Shu~\etal~\cite{shu2019meta} utilize an MLP to explicitly learn the weighting function.
Recently, Bohdal~\etal~\cite{bohdal2021evograd} presented EvoGrad to compute gradients more efficiently by preventing the computation of second-order derivatives in~\cite{shu2019meta}.
However, these methods are targeted for multi-class single-label classification.
Therefore, we present the multi-label meta weight net for predicate classification, with an MLP that output a weight for each class loss. 

\section{Multi-label meta weight network}
\label{sec:method}

Dynamic scene graph generation~\cite{cong2021spatial} takes a video as the input and generates directed graphs whose objects of interest are represented as nodes, and their relationships are represented as edges.
Each relationship edge, along with its connected two object nodes, form a $\langle$subject, predicate, object$\rangle$ semantic triplet.
These directed graphs are structural representations of the video's semantic information.
Highly related to dynamic scene graph generation, video relation detection~\cite{shang2017video} also outputs $\langle$subject, predicate$\rangle$ object triplets, aiming to classify and detect the relationship between object tubelets occurring within a video.
Due to the high similarity between the two tasks, we consider them both in the experiments.
For brevity, in this paper, we use the term dynamic scene graph generation to denote both tasks throughout this paper.

Action Genome ~\cite{ji2020action} and VidOR~\cite{shang2019annotating} are two popular benchmark datasets for dynamic scene graph generation.
However, both datasets suffer from a long-tailed distribution in predicate occurrences, as shown in Figure~\ref{fig:ag}.
The evaluation metrics forgo the class-wise differences and count all classes during inference, resulting in a trained predicate classifier with a strong bias toward head classes such as in\_front\_of and next\_to.
Although these predicate classes are often spatial-oriented and object-agnostic, tail classes like carrying, twisting, and driving are of more interest to us.
In addition to the long-tailed distribution, predicate classification faces another challenge.
Since multiple relationships can occur between a subject-object pair simultaneously, predicate classification is a multi-label classification problem.
The co-occurrence of labels leads to head-class predicate labels frequently appearing alongside tail-class predicate labels, further exacerbating the imbalance problem.

In this paper, we propose a meta-learning framework that addresses on the long-tailed multi-label predicate classification task. 
We introduce a Multi-Label Meta Weight Net (ML-MWN) to learn a weight vector for each training instance's multi-label loss.
The gradient of the sum of weighted loss is then calculated to optimize the classifier network's parameters during backward propagation.
Our model-agnostic approach can be incorporated into existing dynamic scene graph generation methods.
In particular, the framework includes two stages:
(1) Relation feature extraction, where we use existing dynamic scene graph generation methods to obtain the feature representation of the relation instances, and (2) multi-label meta-weighting learning.
We adopt a meta-learning framework to re-weight each instance's multi-label loss and propose learning an explicit weighting function that maps from training loss to weight vector.
We learn a weight vector for each training instance to re-weight its multi-label loss, \ie multi-label binary cross-entropy loss.
We achieve this by using an MLP, which takes the multi-label training loss as input and outputs the weight vector.
We sample a meta-validation set to guide the training of MLP.
Ideally, the meta-validation set should be clean and free from the long-tailed issue, as in~\cite{shu2019meta}.
However, we cannot sample such a clean meta-validation set due to the label-occurrence issue.
To deal with the issue, we adopt the inverse frequency binary cross-entropy loss on meta-validation set.
In the following sections, we describe the ML-MWN and the meta-learning framework in detail.

\subsection{Learning weights for multi-label losses}
\label{sec:net}
Let $x_i$ denote the feature representation of $i$-th relation instance from the training set $\mathcal{D}$ and $y_i \in \mathbb{R}^C$ represent the corresponding multi-label one-hot vector, where $\mathcal{D} = \{ x_i, y_i \}_{i=1}^N$.
The multi-label predicate classifier network is represented by $f_\theta$ with $\theta$ as its parameters.
To enhance the robustness of training in the presence of long-tailed multi-label training instances, we impose weights 
$w_{i, c}$
on the $i$-th instance's $c$-th class loss $l_{i, c}$.
Instead of pre-specifying the weights based on class size~\cite{DBLP:conf/iccv/LinGGHD17, tian2022striking}, we learn an explicit weighting function directly from the data.
Specifically, we propose the ML-MWN (Multi-Label Meta Weight Net) denoted by $g_\phi$, with $\phi$ as its parameters, to obtain the weighting vector for each relation instance's multi-label loss.
We use the loss from $f_\theta$ as the input.

A small meta-validation set $\mathcal{\widehat{D}} = \{x_j, y_j\}_{j=1}^M$, where $M$ is the number of meta-validation instances and $M \ll N$, is sampled to guide the training of ML-MWN.
The meta-validation set does not overlap with the training set.
The weighted losses are then calculated to guarantee that the learned multi-label predicate classifier is unbiased toward dominant classes.

During training, the optimal classifier parameter $\theta^*$ can be extracted by minimizing the training loss:
\begin{equation}
    L^{train}(\theta) = \frac{1}{n} \frac{1}{C} \sum_{i=1}^{n} \sum_{c=1}^C w_{i, c} \cdot l_{i, c} ,
\end{equation}
where $n$ is the number of training instances in a batch, and $C$ is the number of classes.
During inference, we only use the optimal classifier network $f_{\theta^*}$ for evaluation.

\begin{figure*}
    \centering
    \includegraphics[width=.9\linewidth]{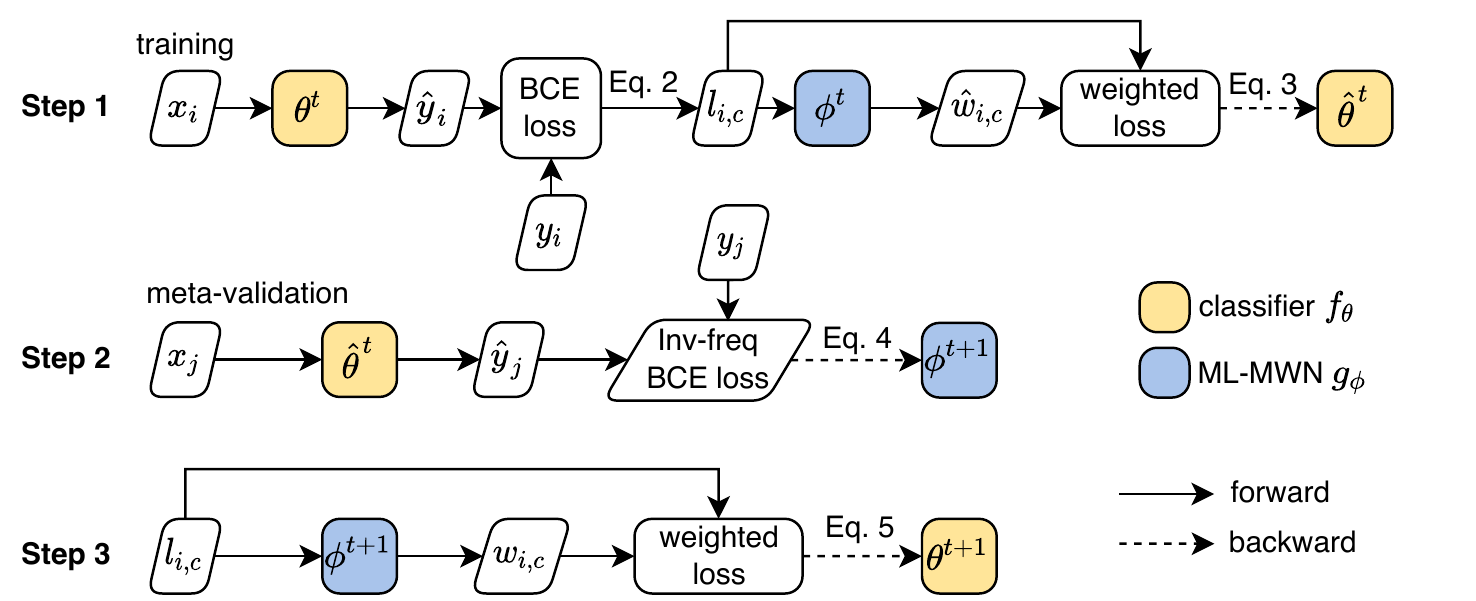}
    \caption{
        An overview of our proposed meta-learning process. We omit the relation feature extraction parts for simplification. $x_i$ and $x_j$ represent the features of the training instance and meta-validation instance, respectively. During a training batch, there are three steps: 1. Calculate the weighted loss and obtain a pseudo classifier; 2. Evaluate the pseudo classifier on the meta-validation set and update the ML-MWN; 3. Compute the new weighted loss with updated ML-MWN then update the classifier. The Binary Cross-Entropy (BCE) loss is adopted for multi-label classification. The inverse-frequency BCE loss is used to simulate an unbiased meta-validation set. During inference, only the predicate multi-label classifier network is utilized.
    }
    \label{fig:overview}
\end{figure*}

\subsection{The meta-learning process}
\label{sec:process}

We adopt a meta-learning framework to update the classifier and ML-MWN.
The meta-validation set represents the unbiased relation instances following a balanced predicate class distribution. 
Due to the multi-label classification label-occurrence issue~\cite{zhang2021deep}, we employ an inverse frequency BCE loss on the meta-validation set to simulate a balanced label distribution.
As illustrated in Figure~\ref{fig:overview}, the process comprises three main steps to optimize $\theta$ and $\phi$ within a batch.

Suppose we are at $t$-th iteration during training.
First, for a batch of $n$ training instances with corresponding feature representations and multi-labels $\{x_i, y_i\}, 1 \leq i \leq n$, we feed $x_i$ into the classifier and obtain $\hat{y}_{i} = f_{\theta^t}(x_i) \in \mathbb{R}^C$.
The unweighted BCE training loss is calculated as
\begin{equation}
\label{eq:train_loss}
    l_{i, c}(\theta^t) = - y_{i,c} \cdot \log \left( \hat{y}_{i, c}(\theta^t) \right) + (1 - y_{i, c}) \cdot \log \left( 1 - \hat{y}_{i, c}(\theta^t) \right).
\end{equation}
Then $l_{i, c}$ is fed into the ML-MWN to obtain the weight $\hat{w}_{i, c} = g_{\phi^t} \left( l_{i, c} (\theta^t) \right) $.
After calculating the weighted loss as $\hat{w}_{i, c} \cdot l_{i, c}$, we update $\theta^t$:
\begin{equation}
\label{eq:pesudo_theta}
    \hat{\theta}^t = \theta^t - \alpha \frac{1}{n}\frac{1}{C} \left. \sum_{i=1}^n \sum_{c=1}^C g'_{\phi^t}(l_{i,c}(\theta^{t})) \nabla_{\theta^t} l_{i,c}(\theta^t) \right |_{\theta^t},
\end{equation}
where $\alpha$ is the step size. We call the updated $\hat{\theta}^t$ the pseudo classifier parameters since they are not used for the next batch.

In the second step, we update the ML-MWN parameters based on the meta-validation loss.
We feed the meta-validation relation instance into the pseudo classifier and obtain $\hat{y}_{j} = f_{\hat{\theta}^t}(x_j) \in \mathbb{R}^C$.
Let $M_c$ denote the total number of relation instances belonging to predicate class $c \in \{1, \dots, C\}$.
The frequency of a predicate class is calculated as $freq(c) = M_c / M$.
By using inverse frequency weighting, the meta-validation loss is re-balanced to mimic a balanced predicate label distribution.
We then update the ML-MWN parameters $\phi$ on the meta-validation data:
\begin{multline}
\label{eq:phi}
\begin{aligned}
    \phi^{t+1} &= \phi^{t} - \beta \left . \frac{1}{M} \sum_{j=1}^M \sum_{c=1}^C \frac{1}{freq(c)} \nabla_{\phi^t} l_{j, c}(\hat{\theta}^{t}) \right |_{\phi^{t}} \\
               &= \phi^{t} - \beta \left . \sum_{c=1}^C \frac{M}{M_c} \nabla_{\phi^t} l_{j, c}(\hat{\theta}^{t}) \right |_{\phi^{t}} ,
\end{aligned}
\end{multline}
where $\beta$ is the step size.

Lastly, the updated $\phi^{t+1}$ is employed to output the new weights $w_{i, c}$.
The new weighted losses are used to improve the parameters $\theta$ of the classifier network:
\begin{equation}
\label{eq:theta}
    \theta^{t+1} = \theta^{t} - \\ \alpha \frac{1}{n}\frac{1}{C} \left . \sum_{i=1}^n \sum_{c=1}^C g'_{\phi^{t+1}}(l_{i,c}(\theta^t)) \nabla_{\theta^t} l_{i,c}(\theta^t) \right |_{\theta^{t}}.
\end{equation}
The ultimate goal is to guide the classifier network to achieve a balanced performance on the unbiased meta-validation set.
The sequences of steps are shown in Algorithm~\ref{alg:meta}.
By alternating between standard and meta-learning, we can learn unbiased dynamic scene graphs by specifically increasing the focus on those examples and predicate classes that do not often occur in a dataset.

\begin{algorithm}[t]
 \caption{The ML-MWN learning algorithm}
\label{alg:meta}
\begin{algorithmic}[1]
 \REQUIRE Training data set $\mathcal{D}$, meta-validation set $\mathcal{\widehat{D}}$, 
 max epochs $N_{Epoch}$
 \ENSURE Predicate multi-label classifier network parameter $\theta^*$
\FOR{t = 1 \TO $N_{Epoch}$ }
    \FOR{each mini batch $\{x_i, y_i\}, 1 \leq i \leq n$}
        \STATE{Calculate the prediction $\hat{y}_i$.}
        \STATE{Calculate the unweighted loss using Eq.~\ref{eq:train_loss}.}
        \STATE{Formulate the pseudo predicate classifier $\hat{\theta}^{t}$ by Eq.~\ref{eq:pesudo_theta}.}
        \STATE{Get meta-validation instances $\{x_j, y_j\} \in \mathcal{\widehat{D}}$.}
        \STATE{Update $\phi^{t+1}$ by Eq.~\ref{eq:phi}.}
        \STATE{Update $\theta^{t+1}$ by Eq.~\ref{eq:theta}.}
    \ENDFOR
\ENDFOR
\end{algorithmic}
\end{algorithm}

\section{Experiments}
\label{sec:exp}

\subsection{Datasets}
\subsubsection{Action Genome}\cite{ji2020action} is a dataset which provides frame-level scene graph labels.
It contains 234,253 annotated frames with 476,229 bounding boxes of 35 object classes (without person) and 1,715,568 instances of 25 relationship classes.
For the 25 relationships, there are three different types: (1) attention relationships indicating if a person is looking at an object or not, (2) spatial relationships describing where objects are relative to one another, and (3) contact relationships denoting the different ways the person is contacting an object.
In AG, there are 135,484 subject-object pairs.
Each pair is labeled with multiple spatial relationships (\eg $\langle$phone-in front of-person$\rangle$ and $\langle$phone-on the side of-person$\rangle$) or contact relationships (\eg $\langle$person-eating-food$\rangle$ and $\langle$person-holding-food$\rangle$).
There are three strategies to generate a scene graph with the inferred relation distribution~\cite{cong2021spatial}:
(a) \emph{with constraint} allows each subject-object pair to have one predicate at most.
(b) \emph{semi constraint} allows a subject-object pair has multiple predicates. The predicate is regarded as positive only if the corresponding confidence is higher than the threshold (0.9 in the experiments).
(c) \emph{no constraint} allows a subject-object pair to have multiple relationships guesses without constraint.

\paragraph{Evaluation metrics.}
We have three tasks for evaluation following~\cite{cong2021spatial}:
(1) predicate classification (PREDCLS): with the subject and object's ground truth labels and bounding boxes, only predict predicate labels of the subject-object pair. 
(2) scene graph classification (SGCLS): with the subject and object's ground truth bounding boxes given, predict the subject, object's label and their corresponding predicate.
(3) scene graph detection (SGDET): detect the subject and object's bounding boxes and predict the subject, object, and predicate's labels.
The object detection is regarded as positive if the IoU between the predicted and ground-truth box is at least 0.5.
Since traditional metrics Recall@K (R@K) are not able to reflect the impact of long-tailed data, we use the mean Recall@K (mR@K), which evaluates the R@K (K = [10, 20, 50] of each relationship class and averages them. We follow the same selection of K as~\cite{cong2021spatial}.

\paragraph{Implementation details.}
We randomly sample 10\% samples from the training set as the meta-validation set.
In line with~\cite{cong2021spatial}, we adopt the Faster-RCNN~\cite{2015NIPS-faster} based on the ResNet101~\cite{2016CVPR-ResNet} as the object detection backbone.
The Faster-RCNN model is trained on AG and provided by Cong~\etal~\cite{cong2021spatial}.
We use an AdamW~\cite{loshchilov2019adamw} optimizer with an initial learning rate $1e^{-4}$ and batch size 1 to train our relation feature model STTran part.
We train ML-MWN using SGD with a momentum of 0.9, weight decay of 0.01, and an initial learning rate of 0.01. 
We train for 10 epochs. 
Other hyperparameter settings are identical to Cong~\etal~\cite{cong2021spatial}.
If not specified, the ML-MWN is an MLP of 1-100-1.
\\\\
\subsubsection{VidOR}\cite{shang2019annotating} is a dataset that includes 10,000 user-generated videos selected from YFCC-100M \cite{thomee2016yfcc100m}, totaling approximately 84 hours of footage.
It contains 80 object categories and 50 predicate categories.
Besides providing annotated relation triplets, the dataset also provides bounding boxes of objects.
VidOR is split into a training set with 7,000 videos, a validation set with 835 videos, and a testing set with 2,165 videos.
Since the ground truth of the test set is unavailable, we follow~\cite{liu2020beyond,qian2019video, xie2020video, su2020video} and use the training set for training and the validation set for testing.
We report the analysis of method performance on the VidOR validation set.

\begin{table*}[t!]
\centering
\begin{tabular}{lccccccccc}
\toprule
& \multicolumn{3}{c}{PredCLS} & \multicolumn{3}{c}{SGCLS} & \multicolumn{3}{c}{SGDET} \\
\cmidrule(lr){2-4} \cmidrule(lr){5-7} \cmidrule(l){8-10}
& mR@10 & mR@20 & mR@50 & mR@10 & mR@20 & mR@50 & mR@10 & mR@20 & mR@50 \\
\midrule
STTran~\cite{cong2021spatial} & 37.96 & 39.65 & 39.66 & 27.61 & 28.14 & 28.14 & 17.89 & 21.76 & 22.89 \\
STTran + MW-Net~\cite{shu2019meta} & 40.29 & 42.21 & 42.24 & 30.21 & 30.90 & 30.90 & 20.06 & 23.66 & 24.99 \\
\rowcolor{mygray} STTran + ML-MWN & \textbf{43.23} & \textbf{44.43} & \textbf{44.64} & \textbf{32.13} & \textbf{32.70} & \textbf{32.72} & \textbf{23.46} & \textbf{27.13} & \textbf{28.52} \\
\bottomrule
\end{tabular}
\caption{Evaluating the effect of meta learning on Action Genome in the \emph{with constraint} setting. Enriching the recent STTran approach with meta learning improves recall across all metrics, with the best results achieved using the proposed multi-label meta weighting.
}
\label{tab:AG_SOTA_with}
\end{table*}

\begin{table*}
\centering
\begin{tabular}{lccccccccc}
\toprule
 & \multicolumn{3}{c}{PredCLS} & \multicolumn{3}{c}{SGCLS} & \multicolumn{3}{c}{SGDET} \\
\cmidrule(lr){2-4} \cmidrule(lr){5-7} \cmidrule(l){8-10}
& mR@10 & mR@20 & mR@50 & mR@10 & mR@20 & mR@50 & mR@10 & mR@20 & mR@50 \\
\midrule
STTran~\cite{cong2021spatial} & 49.94 & 59.07 & 59.77 & 40.17 & 44.27 & 44.51 & 21.63 & 31.36 & 40.96 \\
STTran + MW-Net~\cite{shu2019meta} & 52.61 & 62.32 & 63.1 & 43.12 & 47.11 & 47.77 & 24.19 & 34.83 & 43.85 \\
\rowcolor{mygray} STTran + ML-MWN & \textbf{55.95} & \textbf{65.79} & \textbf{68.01} & \textbf{46.20} & \textbf{50.60} & \textbf{50.83} & \textbf{26.21} & \textbf{40.12} & \textbf{49.96} \\
\bottomrule
\end{tabular}
\caption{Evaluating the effect of meta learning on Action Genome in the \emph{semi constraint} setting. Similar to the \emph{with constraint} setting, the proposed multi-label meta weighting obtains the best results across all metrics.
}
\label{tab:AG_SOTA_semi}
\end{table*}

\begin{table*}
\centering
\begin{tabular}{lccccccccc}
\toprule
 & \multicolumn{3}{c}{PredCLS} & \multicolumn{3}{c}{SGCLS} & \multicolumn{3}{c}{SGDET} \\
\cmidrule(lr){2-4} \cmidrule(lr){5-7} \cmidrule(l){8-10}
& mR@10 & mR@20 & mR@50 & mR@10 & mR@20 & mR@50 & mR@10 & mR@20 & mR@50 \\
\midrule
STTran~\cite{cong2021spatial} & 52.61 & 68.30 & 82.90 & 42.52 & 51.14 & 64.77 & 21.64 & 30.64 & 35.53 \\
STTran + MW-Net~\cite{shu2019meta} & 55.12 & 70.42 & 85.45 & 45.55 & 54.46 & 67.24 & 24.24 & 34.25 & 37.98 \\
\rowcolor{mygray} STTran + ML-MWN & \textbf{57.13} & \textbf{74.22} & \textbf{89.24} & \textbf{48.48} & \textbf{57.65} & \textbf{70.15} & \textbf{27.59} & \textbf{36.67} & \textbf{40.57} \\
\bottomrule
\end{tabular}
\caption{Evaluating the effect of meta learning on Action Genome in the \emph{no constraint} setting. Also in this challenging setting, our approach works best over all metrics. 
}
\label{tab:AG_SOTA_no}
\end{table*}

\paragraph{Evaluation metrics.}
We use the relation detection task for evaluation.
The output requires a $\langle$subject, predicate, object$\rangle$ triplet prediction, along with the subject and object boxes.
We adopt mR@K (K = [50, 100]) as the evaluation metric.
We disregard the mAP used in Chen~\etal~\cite{chen2021social} because we are more concerned with covering ground truth relationships belonging to tail classes during predictions.
\textbf{Calculating mR@K.} 
For annotated video $I_v$, its $G_v$ ground truth relationship triplets contain $G_{v, c}$ ground truth triplets with relationship class $c$.
With $C$ relationship classes, the model successfully predicts $T_{v, c}^K$ triplets.
In the $V$ videos of validation/test dataset, for relationship $c$, there are $V_c$ videos containing at least one ground truth triplet with this relationship.
The R@K of relationship $c$ can be calculated:
\begin{equation}
    R@K_c=\frac{1}{V_c}\sum_{v=1,G_{v, c}\neq0}^{V_c}\frac{T_{v, c}^K}{G_{v, c}}
\end{equation}
Then we can calculate
\begin{equation}
    mR@K=\frac{1}{C}\sum_{c=1}^{C}R@K_c.
\end{equation}

\paragraph{Implementation details.}
We randomly sample 10\% samples from the training set as the meta-validation set.
Our experiments are conducted using 1 NVIDIA V100 GPU.
We adopt the same training strategy of Chen~\etal~\cite{chen2021social} for the relation feature extraction model. 
First, we detect all objects in each video frame using Faster R-CNN~\cite{2015NIPS-faster} with a ResNet-101~\cite{2016CVPR-ResNet} backbone trained on MS-COCO~\cite{lin2014microsoft}.
The detected bounding boxes are linked with the Deep SORT tracker~\cite{2017ICIP-DeepSort} to obtain individual object tubelets. 
Then, each tubelet is paired with any other tubelet to generate the tubelet pairs.
We extract spatial location features~\cite{sun2019video}, language features, I3D features, and location mask features for each pair.
Then the multi-modal features are used as the representation of the relation instance. 
For the classifier and ML-MWN, we use an SGD optimizer with an initial learning rate of 0.01 and train 10 epochs.

\subsection{Multi-label meta weighting on top of the state-of-the-art}
\paragraph{Video scene graph generation.}
First, we investigate the effect of incorporating our meta-learning approach on top of existing state-of-the-art methods for scene graph generation in videos and video relation detection.
We build upon the recent STTran approach of Cong~\etal~\cite{cong2021spatial} for video scene graph generation.
We compare STTran as is and as a baseline that uses conventional meta-learning without considering the multi-label nature of scene graphs, namely MW-Net~\cite{shu2019meta}.
Table~\ref{tab:AG_SOTA_with} shows the results for the \emph{with constraints} setting.
Across the PredCLS, SGCLS, and SGDET tasks, incorporating our meta-learning approach improves the results.
For PredCLS, our proposed STTran + ML-MWN enhances mR@10 by \textbf{5.27}, compared to the STTran baseline.
On mean recall @ 50, we improve the scores by \textbf{4.98}, from 39.66 to 44.64.
On SGDET, the mean recall @ 50 increases from 22.89 to 28.52.
The MW-Net baseline already improves the STTran results, emphasizing the overall potential of meta-learning to address the long-tailed nature of scene graphs.
However, our proposed multi-label meta-learning framework performs best across all tasks and recall thresholds.
This improvement is a direct result of increasing the weight of classes in the long tail when optimizing the classifier network.

The results are consistent for the \emph{semi constraint} and \emph{no constraint} settings, as shown in Table~\ref{tab:AG_SOTA_semi} and Table~\ref{tab:AG_SOTA_no}.
In Table~\ref{tab:AG_SOTA_semi}, the mean recall is higher than in the \emph{with constraint} setting since more predicted results are involved.
For the SGCLS task, our framework achieves 50.60\% on mR@20, which is 6.33\% better than STTran and 3.49\% better than STTran + MW-Net.
Our framework outperforms all metrics in the \emph{no constraint} setting.
In particular, for SGDET, our method reaches 27.59\% at mR@10, 5.95\% better than STTran, and 3.35\% higher than STTran + MV-Net. We conclude that our meta learning framework is effective for video scene graph generation and can be adopted by any existing work.
In Table~\ref{tab:AG_SOTA_no}, the mean recall is the highest among the three settings. Unlimited predictions contribute to enhanced recall performance.
Under this setting, STTran + ML-MWN still achieves the best on all metrics across all tasks.
The results prove our method's generality on various tasks with different settings.

\begin{figure*}
    \centering
    \includegraphics[width=.6\linewidth]{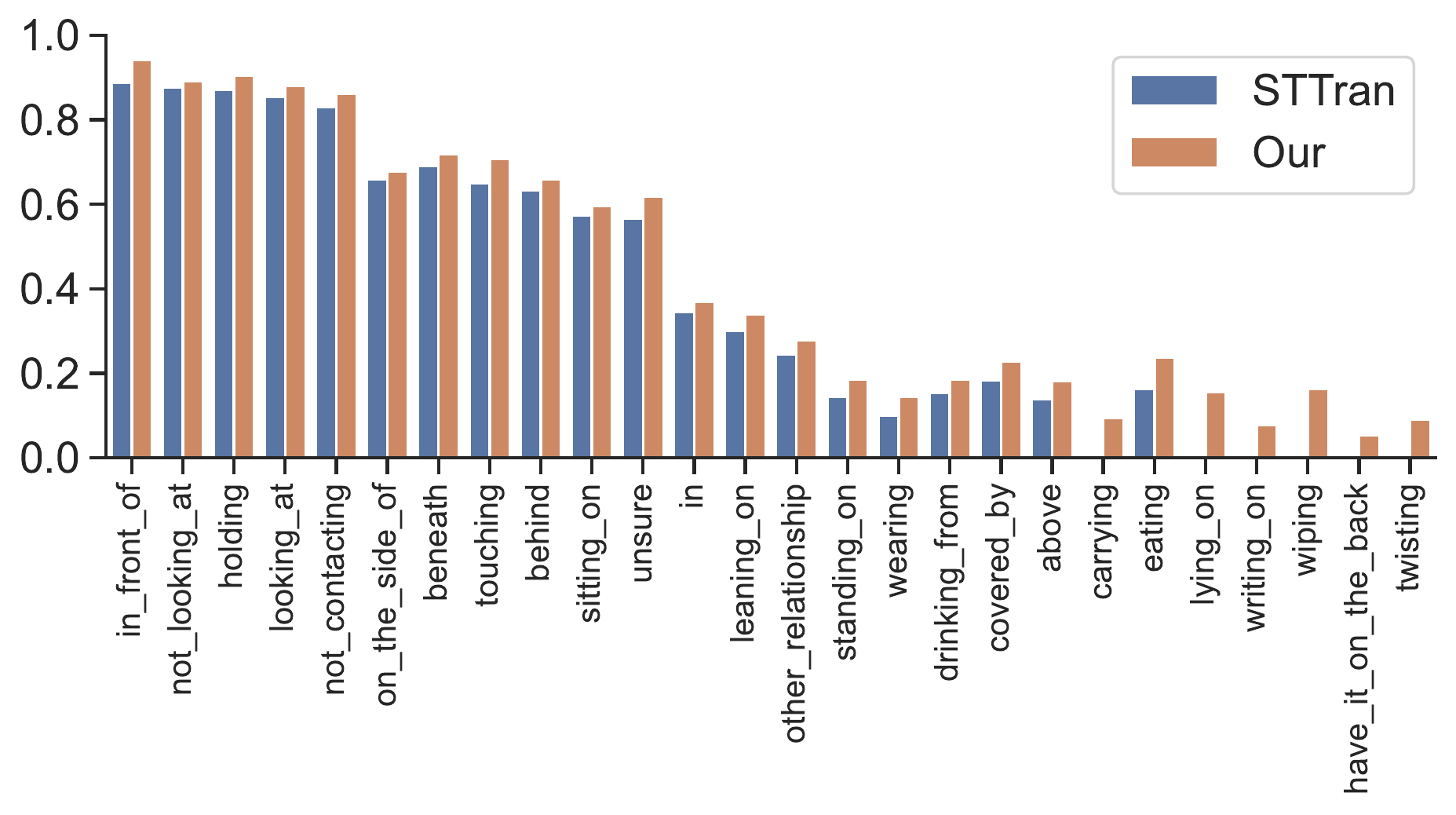}
        \captionof{figure}{Class-wise R@10 comparison of PredCLS on AG. Our framework outperforms STTran on all predicate classes.}
        \label{fig:rel_rec}
        \vspace{0em}
\end{figure*}

\paragraph{Video relation detection.} For video relation detection, we begin with the recent Social Fabric approach by Chen \etal~\cite{chen2021social}.
Table~\ref{tab:VidOR_SOTA} demonstrates the effect of incorporating our proposed meta learning framework for relation detection.
The Social Fabric baseline, which is the current state-of-the-art in this setting, struggles to achieve good results for relation detection using mean recall as metrics.
This underlines the problem's difficulty.
This holds similarly for the baseline by Sun etal~\cite{sun2019video}.
When incorporating MW-Net~\cite{shu2019meta}, the results noticeably improve and further enhance with multi-label meta weighting.
For mR@50, adding our meta-learning on top of Social Fabric boosts the results from 2.37 to 6.35.
We conclude that multi-label meta-learning is crucial in video relation detection to achieve meaningful relation detection recalls across all classes.

\begin{table}
\centering
    \begin{tabular}{lcc}
        \toprule
        \multirow{2}{*}{Method} & \multicolumn{2}{c}{Relation detection} \\
        \cmidrule(){2-3}
        & mR@50 & mR@100 \\
        \midrule
        Sun~\etal~\cite{sun2019video} & 1.48 & 2.78 \\
        Social Fabric (SF)~\cite{chen2021social} & 2.37 & 3.79 \\
        SF + MW-Net~\cite{shu2019meta} & 4.45 & 5.35 \\
        \rowcolor{mygray} SF + ML-MWN & \textbf{6.35} & \textbf{7.54}\\
        \bottomrule
    \end{tabular}
    \captionof{table}{Comparison on the VidOR dataset. Our meta learning framework provides clear improvements for relation detection on top of Social Fabric~\cite{chen2021social}.}
    \label{tab:VidOR_SOTA}
    \vspace{-2em}
\end{table}

\subsection{Analyses, ablations, and qualitative examples}

\begin{table}
    \centering
    \begin{tabular}{lccc}
        \toprule
        \multirow{2}{*}{Architecture} & \multicolumn{3}{c}{PredCLS} \\
        \cmidrule(){2-4}
        & mR@10 & mR@50 & mR@100 \\
        \midrule
        $C$-50-$C$ & 41.34 & 42.24 & 42.56 \\
        \rowcolor{mygray} $C$-100-$C$ & \textbf{43.23} & \textbf{44.43} & \textbf{44.64} \\
        $C$-200-$C$ & 42.16 & 42.85 & 42.97 \\
        $C$-100-100-$C$ & 43.01 & 43.96 & 44.03  \\
        $C$-10-10-$C$ & 42.18 & 42.74 & 42.96 \\
        $C$-10-10-10-$C$ & 42.85 & 44.01 & 44.28  \\
        \bottomrule
        \end{tabular}
        \caption{Performance on AG with constraint for different MLP architecture. The 1-100-1 architecture is the best.}
         \label{tab:MLP}
         \vspace{-2em}
\end{table}

\paragraph{Predicate-level analysis.} We present the class-wise R@10 of the predicate classification task on Action Genome in Figure~\ref{fig:rel_rec}.
Observing Figure~\ref{fig:rel_rec}, we see that our method surpasses STTran~\cite{cong2021spatial} in all predicate categories.
The improvement is much more significant for tail classes with limited training samples compared to head classes.
The superior performance demonstrates that the meta-validation set effectively guides the classifier to balance the tail classes without compromising the performance of head predicate classes.

\paragraph{Ablating the MLP architecture.}
We conduct an ablation study on the MLP architecture for the PredCLS task on Action Genome.
Table~\ref{tab:MLP} shows the results for six structures with varying depths and widths.
We find that maximum width and depth are not necessary, with the best results achieved by the 1-100-1 variant, which we use as default in all experiments.

\begin{figure*}[htb!]
    \centering
    \includegraphics[width=.875\textwidth]{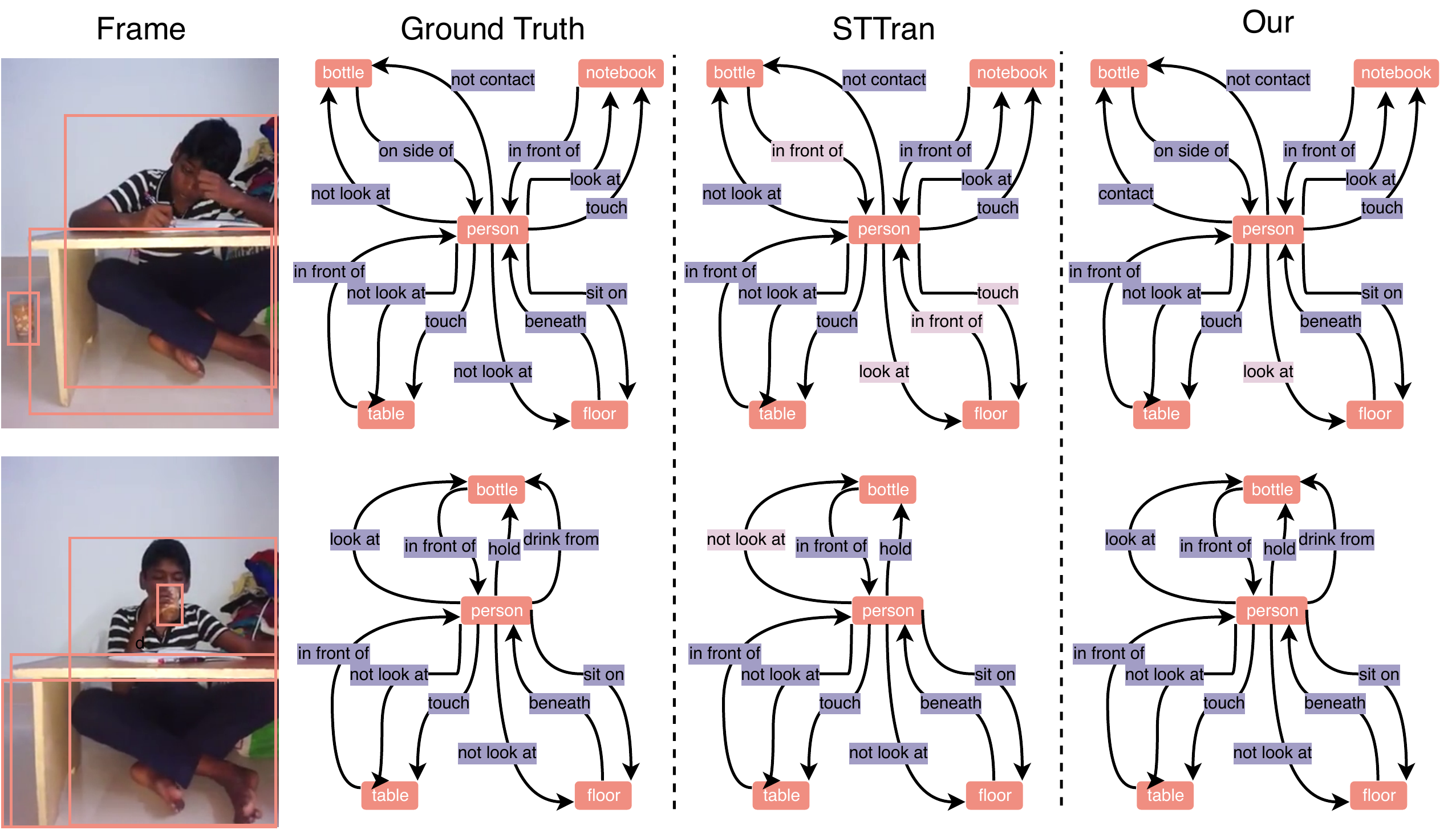}
    \caption{Qualitative comparison on Action Genome predicate classification task. The gray box is the wrongly recognized predicates. Our method performs better than STTran~\cite{cong2021spatial} on recognizing the tail and the head both. }
    \label{fig:example}
    \vspace{0em}
\end{figure*}

\begin{figure*}[htb!]
    \centering
    \includegraphics[width=.875\textwidth]{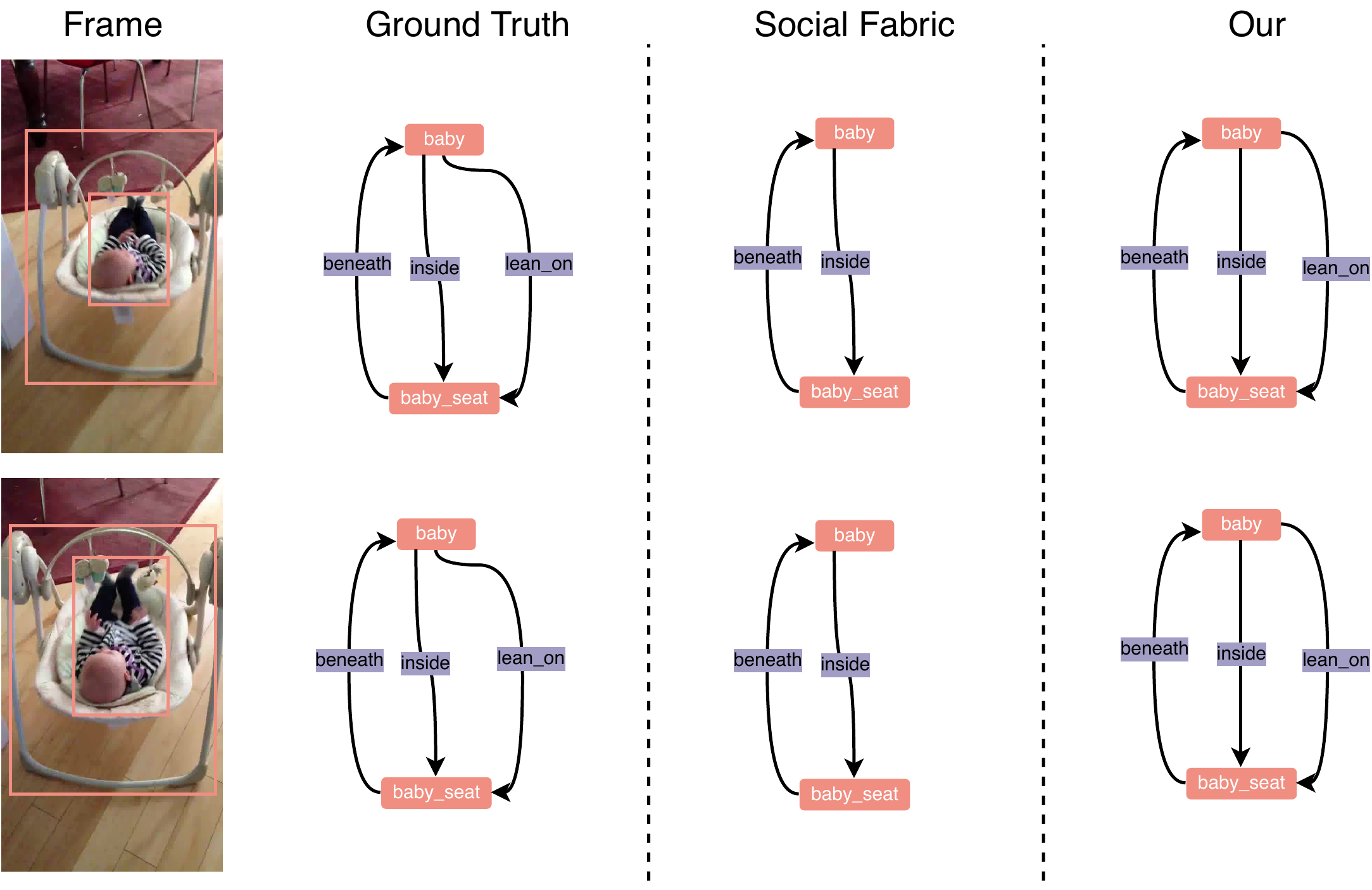}
    \caption{Qualitative comparison on VidOR predicate classification. The Social Fabric baseline \cite{chen2021social} misses the \emph{lean\_on} predicate, while our method detects it correctly. }
    \label{fig:example_vidor}
    \vspace{-1em}
\end{figure*}

\paragraph{Qualitative examples.}
We provide the qualitative results in Figure~\ref{fig:example} and Figure~\ref{fig:example_vidor}.
In Figure~\ref{fig:example}, we compare our method with STTran~\cite{cong2021spatial} on the Action Genome dataset.
Our method demonstrates better recognition of tail predicates in Action Genome. 
In the top row, STTran incorrectly classifies the tail class \emph{beneath} as the head class \emph{in front of}, and \emph{sit on} as \emph{touch}.
In the bottom row, STTran misses \emph{drink from} amongst others, while our method classifies them all correctly.
In Figure~\ref{fig:example_vidor}, we compare our method with Social Fabric~\cite{chen2021social} on the VidOR dataset.
Social Fabric fails to detect the tail class \emph{lean\_on} in all frames, while our method successfully predicts it.

\section{Conclusion}
Predicate recognition plays a crucial role in contemporary dynamic scene graph generation methods, but the long-tailed and multi-label nature of the predicate distribution is commonly ignored. 
We observe that rare predicates on popular benchmarks are inadequately recovered or even disregarded by recent methods. 
To move toward unbiased scene graph generation in videos, we propose a multi-label meta-learning framework that learns to weight samples and classes to optimize any predicate classifier effectively. 
Our approach is versatile and can be incorporated into any existing methods.
Experiments demonstrate the potential of our multi-label meta-learning framework, with superior overall performance and an improved focus on rare predicates. 
We believe our method could be extended to other multi-label long-tailed recognition tasks and may offer inspiration for future research.

\bibliographystyle{ACM-Reference-Format}
\bibliography{long-bib}

\end{document}